\documentclass{article} %
\PassOptionsToPackage{table}{xcolor}
\usepackage{iclr2027_conference,times}

\usepackage{amsmath,amsfonts,bm}

\def\eqref#1{equation~\ref{#1}}

\def\1{\bm{1}}

\DeclareMathAlphabet{\mathsfit}{\encodingdefault}{\sfdefault}{m}{sl}
\SetMathAlphabet{\mathsfit}{bold}{\encodingdefault}{\sfdefault}{bx}{n}

\usepackage{hyperref}
\usepackage{url}
\usepackage{graphicx}
\usepackage[utf8]{inputenc} %
\usepackage[T1]{fontenc}    %
\usepackage{booktabs}       %
\usepackage{multirow}
\usepackage{nicefrac}       %
\usepackage{microtype}      %
\usepackage{xcolor}         %
\usepackage{wrapfig}
\usepackage{makecell}
\usepackage{colortbl}
\usepackage{amsmath}
\usepackage{longtable}
\usepackage{array}
\usepackage{amssymb}
\usepackage[linesnumbered,ruled,vlined]{algorithm2e}
\usepackage{appendix}
\usepackage{placeins}

\usepackage{tabularx} %
\title{Direct Experience World-Model Optimization: Learning the World Beyond Action Imitation}

\author{
\textbf{Xiangcheng Zhan\textsuperscript{1}}\quad
\textbf{Zirui Chen\textsuperscript{2}}\quad
\textbf{Yicheng Zhao\textsuperscript{3}}\quad
\textbf{Ziteng Gao\textsuperscript{1}}\quad
\textbf{Shuo Yang\textsuperscript{1}}\\[6pt]
\normalfont\small\textsuperscript{1}\,Harbin Institute of Technology\\
\normalfont\small\textsuperscript{2}\,Dalian University of Technology\\
\normalfont\small\textsuperscript{3}\,Southern University of Science and Technology
}

\iclrfinalcopy
\hypersetup{
  hidelinks,
  pdftitle={Direct Experience World-Model Optimization: Learning the World Beyond Action Imitation},
  pdfauthor={Xiangcheng Zhan, Zirui Chen, Yicheng Zhao, Ziteng Gao, Shuo Yang}
}

\begin{document}

\maketitle
\fancyhead{}
\renewcommand{\headrulewidth}{0pt}

\begin{abstract}
World-Action Models (WAMs) couple action generation with predictions of how physical interactions unfold. 
However, current post-deployment learning paradigms typically improve behavior without requiring better world predictions. 
Especially in dexterous manipulation, small execution errors can compound in high-dimensional action spaces, hindering policy improvement and pushing interactions beyond the world model's training distribution.
Motivated by this, we propose \textbf{Direct Experience World-Model Optimization (DEWO)}, a post-deployment learning paradigm for WAMs that, alongside action imitation, refines world representations through visual experience to better condition action generation.
Specifically, it identifies interaction turning points and learns from successful and failed futures to support classifier-free guidance. An additional value head estimates task progress from video representations and activates guidance when progress stalls during inference.
Across five DexJoCo tasks, DEWO improves average success across all three WAM formulations. Ablations show that visual supervision from successful and failed continuations improves both prediction and control beyond action supervision alone. On four real-world tasks across Wuji and Sharpa, $3\times3$ grid evaluations show that two rounds of deployment learning increase success from 51.0\% to 71.7\% in cells with at least one initial success, a gain of 20.7 percentage points. These findings support continued predictive learning for improving control through deployment experience, making world modeling an active part of WAM adaptation.

\end{abstract}

\section{Introduction}
\label{sec:intro}

\begin{figure}[t]
    \centering
    \includegraphics[width=0.9\linewidth]{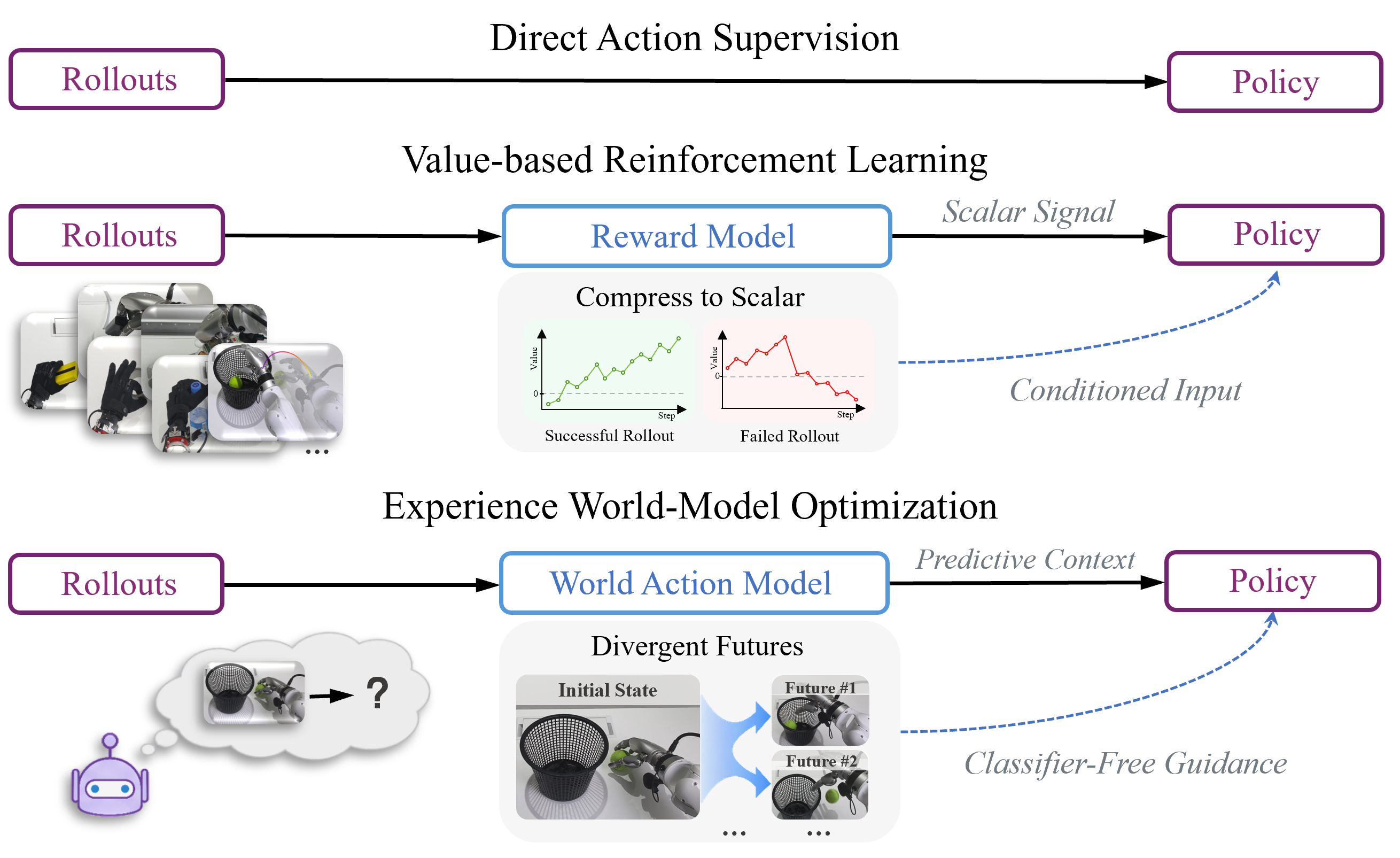}
    \vspace{-4pt}
    \caption{\textbf{Learning signals for post-deployment policy improvement.} Direct action supervision (\textit{top}) learns from action targets. Value-based reinforcement learning (\textit{middle}) uses value estimates to improve the policy. DEWO (\textit{bottom}) learns from observed visual futures and uses the learned success condition to guide actions.}
    \vspace{-4pt}
    \label{fig:pdt}
\end{figure}

World-Action Models (WAMs) couple action generation with predictions of how physical interactions unfold~\citep{tian2025predictive,ye2026world,zhang2026nativevideoactionpretraininggeneralizable}. Visual prediction provides dense supervision for interaction dynamics and can strengthen control even without explicit future generation at execution time~\citep{Finn2016,yuan2026fastwam}. Deployment offers an opportunity to extend this predictive learning to the interactions a robot actually encounters. Yet post-deployment learning through direct action supervision~\citep{ross2011reduction,liu2023robot} or value-based reinforcement learning~\citep{intelligence2025pi,wagenmaker2025steering,yu2026rlinf} primarily targets behavior, without requiring better predictions of deployment interactions (Figure~\ref{fig:pdt}).

This gap is especially relevant in dexterous manipulation, where small action deviations can alter contact and send otherwise similar executions toward different physical futures~\citep{yang2026lamp,feng2026learningdexterousmanipulationquantized}. After such interaction turning points, later actions encounter different object configurations, allowing errors to compound as execution departs from the training distribution~\citep{ross2011reduction}. For a WAM, this creates a coupled adaptation problem: unfamiliar interactions challenge both action generation and the predictive representations that support it. 

These interaction turning points provide the key supervision for this adaptation. A failed continuation from such a context records a valid physical future even when its actions should not be imitated. Paired with a successful continuation from a comparable context, it reveals how similar interactions can evolve differently.
Building on online world-model updates~\citep{qian2026wamrl} and learning from failed interactions~\citep{peng2026fact}, we ask: \emph{how can continued world modeling from matched continuations around interaction turning points improve control beyond action imitation?}

We propose \textbf{Direct Experience World-Model Optimization (DEWO)}, a post-deployment learning paradigm for WAMs that, alongside action imitation, refines world representations through visual experience to better
condition action generation. Observed visual futures refine the predictive representations that support action generation, with action supervision as a complementary signal. Replay exploration identifies interaction turning points and collects matched successful and failed continuations. These continuations reveal how similar interactions can lead to different physical futures. Both outcomes supply visual supervision under their respective outcome conditions, while action targets come only from successful continuations.

At inference, classifier-free guidance~\citep{ho2022classifier,zheng2023guidedflowsgenerativemodeling} steers actions toward the learned success condition. A value head over video representations estimates task progress and activates guidance when it stalls. This connects the learned success condition to selective intervention at moments of insufficient task progress.

On five DexJoCo tasks~\citep{wang2026dexjoco}, DEWO improves average success across three WAM paradigms: joint modeling, video-then-action inverse dynamics, and FastWAM. In matched comparisons, success rises from 80.7\% to 82.3\% on FastWAM and from 64.5\% to 81.2\% on FACT, outperforming SFT on both. Controlled ablations show that visual-future supervision reduces held-out video loss and improves task success beyond action supervision alone. On four real-world tasks across two dexterous-hand platforms, two deployment-learning rounds raise pooled success across all evaluated positions from 20.6\% to 28.9\%. Spatial evaluation shows where these gains occur: success in cells with at least one initial success rises from 51.0\% to 71.7\%.

Our main contributions are:
\begin{itemize}
\item We formulate \textbf{DEWO}, a post-deployment learning paradigm for WAMs centered on continued world modeling from observed visual futures.

\item We develop a framework around interaction turning points that links replay exploration, outcome-conditioned world modeling, and progress-gated guidance. It learns from matched successful and failed futures while reserving action supervision for successful behavior.

\item We demonstrate improved average task success across three WAM paradigms and post-deployment gains on two real-world dexterous-hand platforms. Controlled ablations show that learning from successful and failed visual experience improves both prediction and control beyond action supervision.

\end{itemize}

\section{Direct Experience World-Model Optimization}
\label{sec:method}

\begin{figure}[t]
\centering
\includegraphics[width=\linewidth]{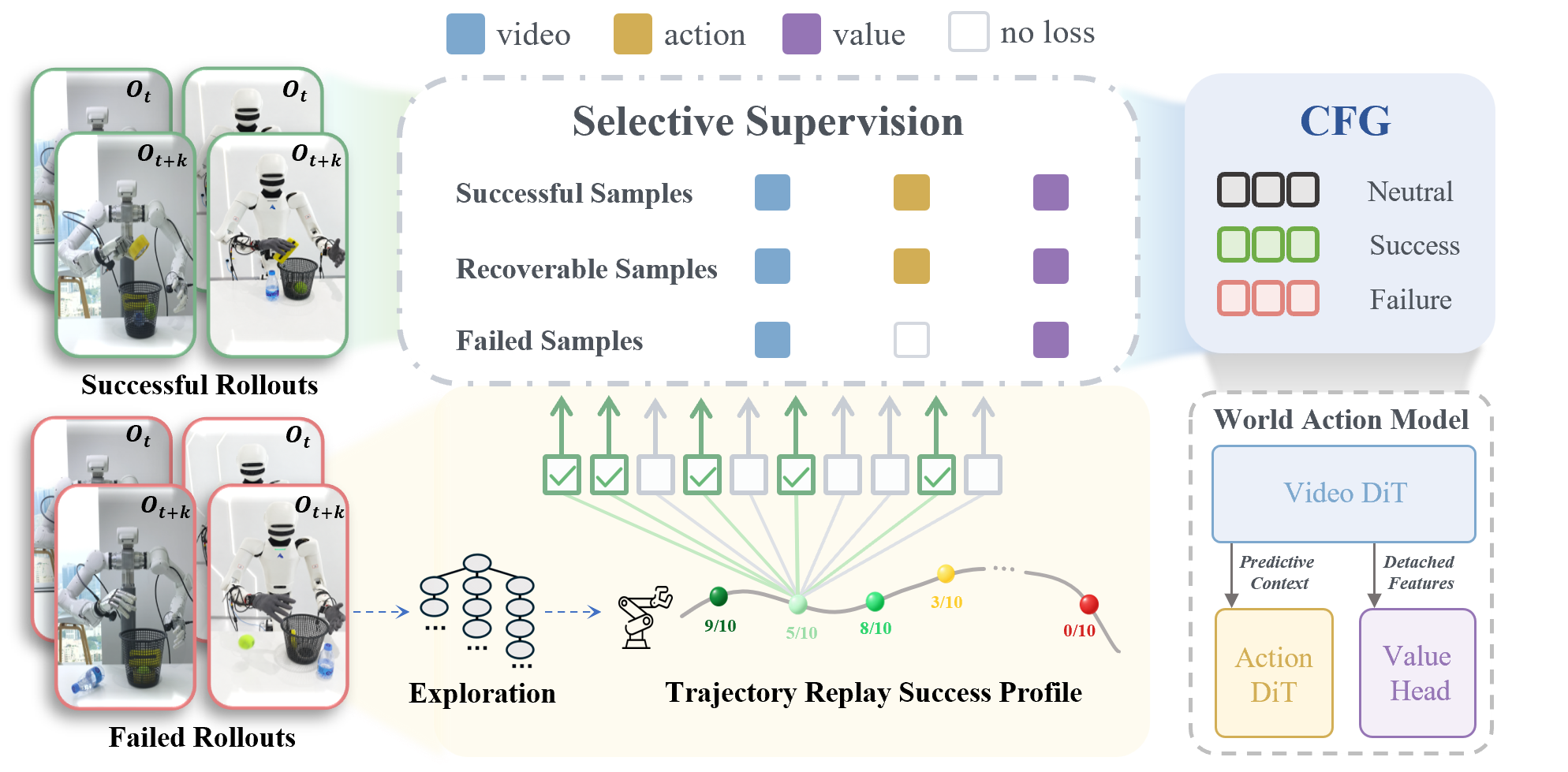}
\caption{\textbf{Overview of Direct Experience World-Model Optimization (DEWO).} Replay exploration locates interaction turning points and collects successful and failed continuations from matched contexts. Their visual futures train outcome-conditioned world modeling, while successful behavior supplies action targets. At deployment, a value head monitors task progress and selectively activates classifier-free guidance using the base and success-conditioned action predictions.
}
\label{fig:overview}
\end{figure}

\subsection{Continued World Modeling from Deployment Experience}
\label{sec:paradigm}

Direct Experience World-Model Optimization (DEWO) formalizes continued world modeling~\citep{kessler2023effectivenessworldmodelscontinual,yao2026sc2wm} as a post-deployment learning paradigm for WAMs. Beyond direct action supervision and value-based reinforcement learning, DEWO retains observed physical futures as learning targets. It focuses on \emph{interaction turning points}, where small deviations can send similar executions toward different futures.

Let $\mathcal R$ denote deployment rollouts and $\mathcal T(\mathcal R)$ their interaction turning points. For each $z\in\mathcal T(\mathcal R)$, $h_z$ denotes the interaction context and $\mathcal C(z)$ the set of observed continuations collected around it. At the paradigm level, DEWO continues world-model learning on these continuations:
\begin{equation}
\theta^{+}
=
\arg\min_{\theta}
\mathbb E_{z\sim\mathcal T(\mathcal R)}
\mathbb E_{\xi\sim\mathcal C(z)}
\left[
\ell_{\mathrm{vis}}(\theta;h_z,c,\xi)
\right],
\label{eq:dewo_paradigm}
\end{equation}
where $c$ is the conditioning input and $\ell_{\mathrm{vis}}$ the WAM's native visual prediction objective. The key distinction is therefore not a new visual loss, but how direct deployment experience is selected and organized for continued predictive learning: around interaction turning points and their subsequent physical futures. Actions from successful experience can additionally provide complementary supervision, which we specify below.

The following sections instantiate this paradigm through turning-point experience collection, outcome-conditioned world-model learning, and progress-gated action guidance (Figure~\ref{fig:overview}).

\subsection{From Interaction Turning Points to Divergent Futures}
\label{sec:critical_experience}

Earlier interaction contexts of a failed rollout may still admit successful alternatives.
DEWO revisits these contexts through replay exploration to identify \emph{interaction turning points} and collect matched continuations that evolve toward different outcomes. 

\paragraph{Replay exploration.}
In simulation, the original failed rollout provides a reference trajectory. At a queried interaction context $h_t$, we restore the corresponding environment state and sample $K$ continuations
$\{\tau_t^{(k)}\}_{k=1}^{K}$ using the deployed checkpoint for the current collection round. Each continuation records its visual evolution, executed actions, and terminal outcome. Let $Y(\tau)\in\{0,1\}$ denote task success. Empirical recoverability is
\begin{equation}
\hat\rho_K(h_t)
=
\frac{1}{K}\sum_{k=1}^{K}Y(\tau_t^{(k)}).
\label{eq:recoverability}
\end{equation}
The original rollout provides a failed continuation anchored at the same interaction context.

\paragraph{Identifying interaction turning points.}
We query contexts progressively along the failed rollout and use changes in empirical recoverability to localize interaction turning points. For successive queries $h_{t_i}$ and $h_{t_{i+1}}$, we detect a boundary when
\begin{equation}
\hat\rho_K(h_{t_i})>0,
\qquad
\hat\rho_K(h_{t_{i+1}})=0.
\label{eq:critical_event}
\end{equation}
At such a boundary, we retain from $h_{t_i}$ the shortest sampled successful continuation together with the corresponding continuation from the original failed rollout. Because both originate from the same restored interaction context, the pair exposes how similar interactions can lead to different subsequent physical futures. We also retain failed windows around sharp intermediate drops in recoverability, which provide additional supervision for unfavorable world evolution. Exact query spacing, retention thresholds, and tie-breaking rules are given in Appendix~\ref{app:critical_experience}. These turning points reflect the current policy and finite sampling budget; zero sampled successes need not imply physical irrecoverability.

\paragraph{Matched experience on physical robots.}
Physical collection follows the same principle without requiring exact state restoration. Human inspection selects candidate interaction contexts from failed rollouts, after which the task-relevant scene and robot configuration are restored and alternative continuations are collected. Matching is defined by comparable task-relevant physical configurations rather than identical images or observation histories. Appendix~\ref{app:critical_experience} describes how these configurations are restored on the robots.

\paragraph{Retained experience.}
The retained successful and failed continuations, $\mathcal D_+$ and $\mathcal D_-$, form our \emph{critical experience}. Both provide visual supervision for how interactions unfold, while only $\mathcal D_+$ provides action targets. These futures support outcome-conditioned world modeling; replay statistics also train a value head for progress estimation during execution (Section~\ref{sec:value_guidance}).

\subsection{Outcome-Conditioned World-Model Learning}

The continuations collected around interaction turning points provide visual evidence of how comparable interaction contexts can evolve toward different outcomes. DEWO uses outcome conditioning to learn these divergent futures while preserving their association with task success or failure. Adding a success or failure label to the original task condition \(c_{\mathrm{base}}\) gives \(c_+\) or \(c_-\).

Successful continuations supervise both visual and action prediction under \(c_+\), whereas failed continuations supervise visual prediction under \(c_-\). The critical-experience objective is

\begin{equation}
\begin{aligned}
\mathcal{L}_{\mathrm{crit}}
={}&\mathbb{E}_{\tau^+\sim\mathcal{D}_+}
\left[
\mathcal{L}_{\mathrm{vis}}(\tau^+\mid c_+)
+\lambda_{\mathrm{act}}\mathcal{L}_{\mathrm{act}}(\tau^+\mid c_+)
\right]\\
&+\mathbb{E}_{\tau^-\sim\mathcal{D}_-}
\left[
\mathcal{L}_{\mathrm{vis}}(\tau^-\mid c_-)
\right].
\end{aligned}
\label{eq:critical_world_model_loss}
\end{equation}

Here, \(\mathcal{L}_{\mathrm{vis}}\) and \(\mathcal{L}_{\mathrm{act}}\) are the native WAM objectives applied to windows from each continuation, and \(\lambda_{\mathrm{act}}\) weights action supervision. Failed continuations thus extend world-model learning beyond the successful behavior available for action supervision. Through the WAM's existing coupling between prediction and action, these predictive updates support action generation. The role-weighted training objective and implementation-specific regularizers are specified in Appendix B.2.

\paragraph{Preparing success-conditioned guidance.} To use the learned success condition during execution, we train a base reference within the same WAM. Alongside the critical experience, we replay retained successful experience \(\mathcal{D}_0\) under \(c_{\mathrm{base}}\) using the native visual and action objectives. We also randomly drop the success label from critical successful examples, replacing \(c_+\) with \(c_{\mathrm{base}}\). Replay provides broader successful task experience, while condition dropout trains the base reference on the same critical continuations used by the success condition. The base and success predictions share model parameters and provide the two action branches for classifier-free guidance. Section 2.4 uses progress estimates to determine when to invoke this guidance.

\subsection{Progress-Gated Action Guidance}
\label{sec:value_guidance}

At inference, we use the learned success condition for classifier-free action guidance. The WAM follows \(c_{\mathrm{base}}\) by default. A value head on its video representations, trained on replay continuations to predict discounted terminal success, monitors progress. Guidance is activated when these estimates fail to increase sufficiently over a recent window of replanning decisions.

At generative time \(s\), let \(f^{\mathrm{act}}_{b,s}\) and \(f^{\mathrm{act}}_{+,s}\) denote the native action predictions under \(c_{\mathrm{base}}\) and \(c_+\), with the current action latent and other inputs shared. With guidance scale \(w\) and progress gate \(g_j\in\{0,1\}\), the guided action prediction at replan \(j\) is

\begin{equation}
f^{\mathrm{act}}_{\mathrm{guided},s}
=
f^{\mathrm{act}}_{b,s}
+
g_jw
\left(
f^{\mathrm{act}}_{+,s}
-
f^{\mathrm{act}}_{b,s}
\right).
\label{eq:progress_gated_action_prediction}
\end{equation}

For flow-matching WAMs~\citep{lipman2023flowmatchinggenerativemodeling}, \(f\) is the predicted velocity field. The gate selectively adds the learned success direction, while visual prediction remains base-conditioned. Appendices~\ref{app:value_guidance} and~\ref{app:dewo_discount_gate} specify value targets, feature extraction, and the gating rule; Appendix~\ref{app:dewo_reference_guidance} analyzes the guidance direction.

\section{Experiments}
\label{sec:dewo_experiments}

We evaluate DEWO's post-deployment gains across prediction--action formulations and track performance over repeated real-world deployment. We also include a complementary construction setting, DEWO-S, which starts from pretrained video and action components. Section~\ref{sec:analysis} analyzes visual supervision, experience selection, and guidance timing.

\subsection{Experimental Setup}
\label{sec:dewo_setup}

\paragraph{Simulation.}
We use five DexJoCo tasks~\citep{wang2026dexjoco}: Water Plant,
Fold Glasses, Hammer Nail, Pick Bucket, and Pinch Tongs. All simulation tests,
including ablations, evaluate each checkpoint on three 50-scene sets per task;
overall averages weight tasks equally. Methods within each adaptation comparison
share evaluation scenes. Collection and evaluation use distinct spatial configurations;
Appendix~\ref{app:simulation_protocol} details the shared evaluation protocol.

\paragraph{Models and comparisons.}
We evaluate two WAMs, FastWAM~\citep{yuan2026fastwam} and FACT~\citep{peng2026fact}, and the VLA $\pi_{0.5}$~\citep{intelligence2025pi05visionlanguageactionmodelopenworld}.
On both WAMs, we compare DEWO with supervised fine-tuning (SFT).
The WAM SFT baseline follows FACT's failure-aware training
scheme, using all rollouts for video
and value/progress supervision, while restricting action imitation
to successful rollouts. For $\pi_{0.5}$, SFT uses only successful
rollouts. We additionally compare RECAP~\citep{intelligence2025pi}
and DSRL~\citep{wagenmaker2025steering} on $\pi_{0.5}$ and FastWAM.
Methods share an initial checkpoint within each base-model comparison;
the $\pi_{0.5}$ and FastWAM comparisons match training-data quantities
and budgets. Implementation details are provided in
Appendix~\ref{app:experimental_protocols}.

Table~\ref{tab:dewo_main}B examines prediction--action coupling within a
shared mixture-of-transformers (MoT) built from Wan2.2 VideoDiT~\citep{wan2025} and
ActionDiT components. The three formulations---Joint, IDM, and FastWAM---are
each evaluated before and after DEWO. Table~\ref{tab:dewo_main}C reports
DEWO-S construction: a Joint WAM is trained from pretrained video and
action components under the DEWO objective and evaluated before further
deployment adaptation.

\paragraph{Real-world evaluation.}
We evaluate grasp-and-place of a water bottle, tape, an eraser, and a
tennis ball with Wuji and Sharpa hands. Success requires grasping the object
and placing it in the target basket. Each object--hand pair uses a
$3\times3$ position grid with ten trials per cell. R0 is the initial
DEWO-S model trained on data collected near the center; R1 and R2 each follow a DEWO
collection--optimization cycle, giving 2,160 trials across eight pairs
and three rounds. We report full-grid success and a fixed subset of
cells with at least one R0 success, including centers. $\pi_{0.5}$ is a
separate reference. Protocols and grid counts appear in
Appendices~\ref{app:real_protocol} and~\ref{app:real_spatial}.

\begin{table}[t]
\centering
\caption{Simulation success rates (\%) over three 50-trial sets per task;
Avg. weights tasks equally. A: matched adaptation comparisons.
B: prediction--action formulation ablation; FastWAM repeats A.
C: DEWO-S construction from pretrained components (\textemdash: no subsequent
adaptation). Per-set results are in Appendix~\ref{app:eval_rounds}.}
\label{tab:dewo_main}
\begingroup
\small
\setlength{\tabcolsep}{3pt}
\renewcommand{\arraystretch}{1.04}
\begin{tabular*}{\linewidth}{@{\extracolsep{\fill}}llcccccc@{}}
\toprule
Model & Method & Water & Fold & Hammer & Pick & Pinch & Avg. \\
 & & Plant & Glasses & Nail & Bucket & Tongs & \\
\midrule
\multicolumn{8}{@{}l}{\textit{A. Matched post-deployment comparisons}} \\
\addlinespace[2pt]
$\pi_{0.5}$ & Initial & 73.3 & 58.0 & 74.7 & 78.7 & 62.7 & 69.5 \\
 & + SFT& 75.3 & 59.3 & 76.7 & 85.3 & 22.7 & 63.9 \\
 & + RECAP & 75.3 & 51.3 & 80.0 & 84.0 & 57.3 & 69.6 \\
 & + DSRL & 76.7 & 54.7 & 81.3 & 86.0 & 22.7 & 64.3 \\
\addlinespace[3pt]
FACT & Initial & 66.0 & 68.0 & 25.3 & 85.3 & 78.0 & 64.5 \\
& + SFT & 68.7 & 69.3 & 41.3 & 86.7 & 84.7 & 70.1 \\
& + DEWO & 94.7 & 78.7 & 44.7 & 92.7 & 95.3 & 81.2 \\
FastWAM & Initial & 88.7 & 72.0 & 74.7 & 92.0 & 76.0 & 80.7 \\
 & + SFT & 73.3 & 74.7 &  78.7 & 84.7 & 69.3 & 76.1 \\
 & + RECAP & 78.0 & 72.7 & 71.3 & 90.7 & 71.3 & 76.8 \\
 & + DSRL & 73.3 &  78.0 & 74.7 & 90.7 & 66.0 & 76.5 \\
 & + DEWO  &  90.7 & 77.3 & 72.7 &  93.3 &  77.3 & 82.3 \\
\midrule
\multicolumn{8}{@{}l}{\textit{B. Prediction--action formulation ablation}} \\
\addlinespace[2pt]
IDM & Initial & 86.0 & 68.0 &  \textbf{85.3} & 86.0 & 85.3 & 82.1 \\
 & + DEWO & \textbf{96.7} & 72.0 & 81.3 & 88.7 & 81.3 & 84.0 \\
\addlinespace[3pt]
FastWAM & Initial & 88.7 & 72.0 & 74.7 & 92.0 & 76.0 & 80.7 \\
 & + DEWO & 90.7 &  77.3 & 72.7 & 93.3 & 77.3 & 82.3 \\
\addlinespace[3pt]
Joint & Initial & 81.3 & 74.7 & 82.0 & 92.7 & 78.7 & 81.9 \\
 & + DEWO & 92.0 & 75.3 & 79.3 & 94.7 & 92.0 & 86.7 \\
\midrule
\multicolumn{8}{@{}l}{\textit{C. Construction from pretrained components}} \\
\addlinespace[2pt]
DEWO-S & \textemdash & 94.7 & \textbf{82.7} & 76.7 & \textbf{97.3} & \textbf{98.0} & \textbf{89.9} \\
\bottomrule
\end{tabular*}
\endgroup
\end{table}

\subsection{Simulation Results}
\label{sec:dewo_simulation}

\paragraph{Post-deployment adaptation in dexterous control.}
Post-deployment updates do not consistently improve dexterous policies
(Table~\ref{tab:dewo_main}A). On FastWAM, SFT, RECAP, and DSRL reach
76.1--76.8\% average success, below the initial 80.7\%.
We attribute this difficulty in part to high-dimensional, contact-sensitive
hand control, where small action changes can disrupt coordinated contacts
and compound during execution~\citep{yang2026lamp}.
DEWO raises FastWAM to 82.3\% and FACT from 64.5\% to 81.2\%, compared
with 70.1\% for FACT SFT. Since WAM SFT also uses visual futures, the
advantage on both models extends beyond simply retaining visual supervision.
Section~\ref{sec:dewo_real_world} examines how deployment gains relate to
the model's initial spatial generalization.

\paragraph{DEWO benefits from joint modeling.}
DEWO improves all three prediction--action formulations, with average gains
of 4.8 percentage points for Joint, 1.9 for IDM, and 1.6 for FastWAM
(Table~\ref{tab:dewo_main}B). Joint benefits most; we hypothesize that its
direct coupling of future prediction and action generation allows their
learning to reinforce each other. This motivates introducing DEWO from the
start of WAM training. DEWO-S applies the DEWO objective while constructing
a Joint MoT from pretrained Wan2.2 VideoDiT and ActionDiT components.
It reaches 89.9\% average success (Table~\ref{tab:dewo_main}C), compared
with 86.7\% for Joint adapted after initial training. These results suggest
that integrating experience-based world-model learning during initial
construction can provide a stronger foundation for subsequent adaptation.
DEWO-S provides the real-world starting point.

\subsection{Real-World Results: Iteration and Spatial Generalization}
\label{sec:dewo_real_world}

\begin{figure}[t]
\centering
{\raggedright\small\sffamily\bfseries (a) Spatial success maps\par}
\vspace{2pt}
\includegraphics[width=\linewidth]{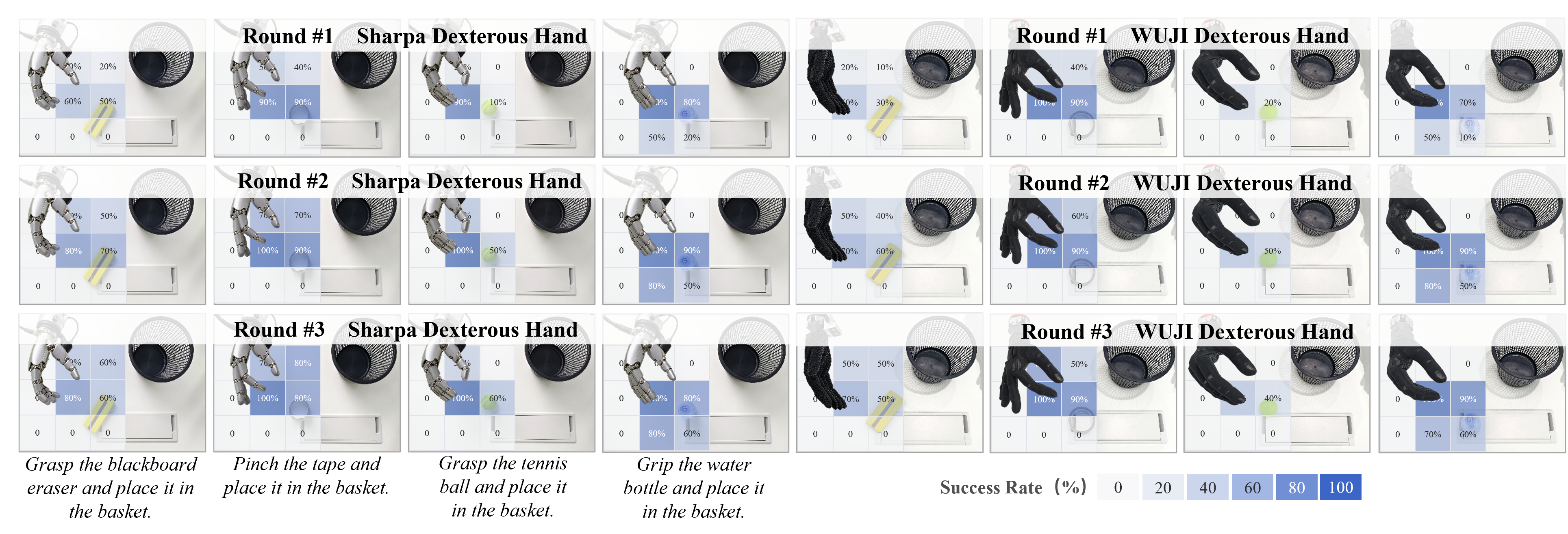}
\par\vspace{5pt}
\noindent\begin{minipage}[t]{0.414\linewidth}
\raggedright\small\sffamily\bfseries (b) Pooled deployment results
\end{minipage}%
\begin{minipage}[t]{0.586\linewidth}
\raggedright\small\sffamily\bfseries (c) Full-grid rates (\%)
\end{minipage}
\par\vspace{2pt}
\includegraphics[width=\linewidth]{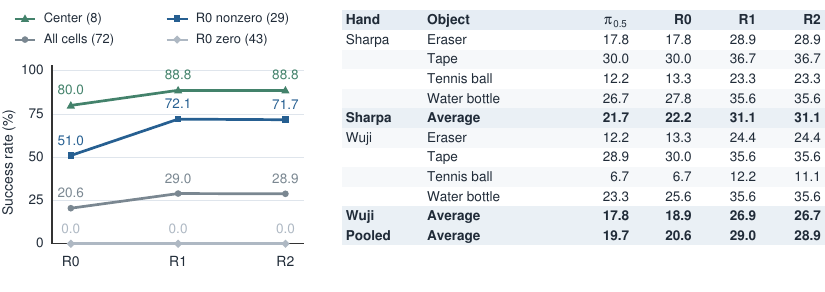}
\caption{\textbf{Spatial and iterative real-world evaluation.}
\textbf{(a)} Two hands, four objects, three rounds; image labels
Round \#1/\#2/\#3 denote R0/R1/R2.
\textbf{(b)} Pooled success for eight centers, 29 R0-nonzero cells
(including centers), 43 R0-zero cells, and all 72 cells. Groups stay
fixed across rounds, with ten trials per cell (720 per round).
\textbf{(c)} Full-grid rates by object and hand; $\pi_{0.5}$ is a separate reference.}
\label{fig:dewo_spatial}
\end{figure}

\paragraph{Deployment learning improves both embodiments.}
All eight object--hand pairs improve from R0 to R2. Pooled full-grid
success rises from 20.6\% (148/720) to 28.9\% (208/720)
(Figure~\ref{fig:dewo_spatial}c). Gains occur on both hands: from 18.9\%
to 26.7\% on Wuji and from 22.2\% to 31.1\% on Sharpa.
Sharpa maintains higher success across rounds. We suspect that its joint
articulation better accommodates the tested grasps, contributing to this
advantage.

\paragraph{Iterative DEWO reinforces initial spatial generalization.}
Comparing R0, R1, and R2 locates the gains from successive updates
(Figure~\ref{fig:dewo_spatial}b). On the fixed 29 cells with at least one
R0 success, pooled success rises from 51.0\% to 72.1\% and 71.7\%.
The improvement concentrates at off-center positions, where success
increases from 40.0\% to 65.7\% and 65.2\%, compared with 80.0\%,
88.8\%, and 88.8\% at the centers. These off-center positions already
admit occasional R0 success despite initial training near the center.
All 43 initially zero cells remain at zero. This pattern indicates that
DEWO primarily makes the initial model's spatial generalization more
reliable: most of the improvement emerges in R1 and is retained in R2.
Appendix~\ref{app:spatial_aggregation} gives the counts and aggregation.

\section{Analysis}
\label{sec:analysis}
\label{sec:dewo_analysis}

Figure~\ref{fig:dewo_components} analyzes consequence supervision,
experience selection, and guidance timing on Water Plant and Fold Glasses.
Training ablations independently adapt the same FastWAM checkpoint on the
same rollout pool; guidance
ablations vary only inference settings on one adapted checkpoint.
Appendix~\ref{app:ablation_protocol} specifies the protocols and offline metric.

\subsection{Learning Consequences Beyond Action Imitation}
\label{sec:dewo_consequences}
\label{sec:dewo_components}

\begin{figure}[t]
\centering
\includegraphics[width=\linewidth]{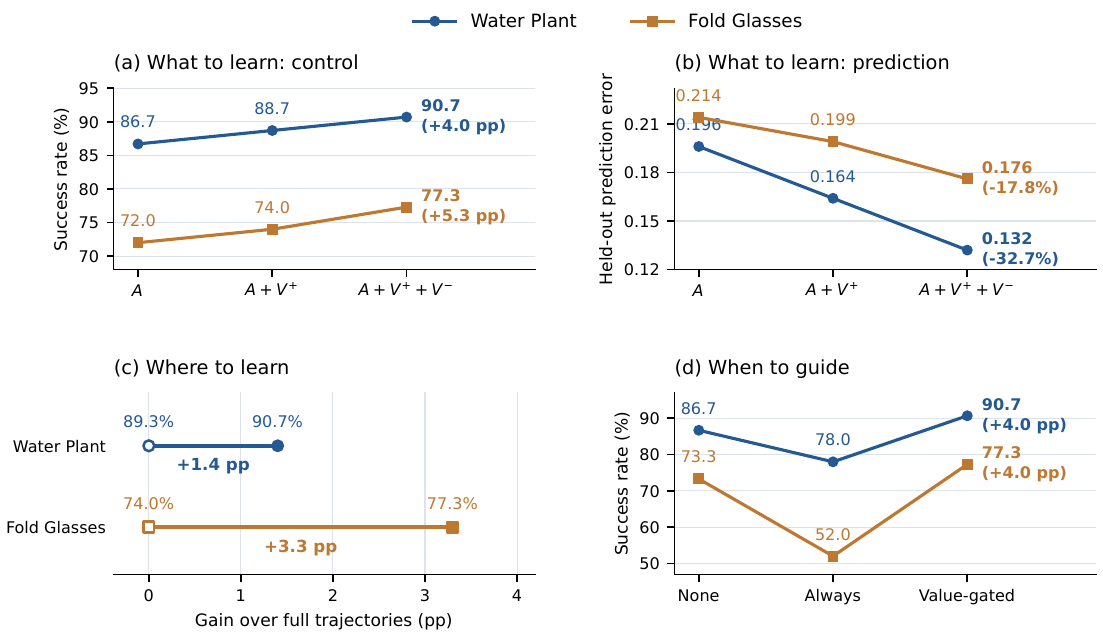}
\caption{\textbf{Component analysis on Water Plant and Fold Glasses.}
$A$: successful-action supervision; $V^+$/$V^-$: successful/failed visual futures.
\textbf{(a,b)} Task success and validation video loss (lower is better),
with changes from $A$. Failed actions are not imitation targets.
\textbf{(c)} Full trajectories (open) versus critical experience (filled),
matching data sizes and training budgets; labels show success.
\textbf{(d)} No guidance, continuous guidance, and value-gated guidance;
gains are relative to no guidance in percentage points (pp).}
\label{fig:dewo_components}
\end{figure}

\paragraph{Consequence supervision improves task success.}
We retain successful-action supervision and add visual targets in two
steps, separating the contribution of successful futures from that of
failed futures.
Adding successful ($V^+$) and then failed ($V^-$) visual futures to
successful-action supervision ($A$) raises success from 86.7\% to 88.7\%
to 90.7\% on Water Plant and from 72.0\% to 74.0\% to 77.3\% on Fold
Glasses (Figure~\ref{fig:dewo_components}a). The second increment comes
from continuations that provide no action-imitation targets, showing that
their observed consequences can improve control through predictive learning.

\paragraph{Predictive improvements accompany the control gains.}
To examine whether the same updates also improve world modeling, we
evaluate the variants using held-out video prediction loss.
This loss falls on both tasks as visual supervision expands from
successful to failed continuations
(Figure~\ref{fig:dewo_components}b). Relative to action-only training, full
visual supervision reduces this loss by 32.7\% on Water Plant and 17.8\%
on Fold Glasses. These complementary measurements show that predictive
adaptation accompanies the control improvements, supporting continued
world-model learning alongside successful-action supervision.

\subsection{Experience Selection and Selective Guidance}
\label{sec:dewo_selection_guidance}

\paragraph{Learning at interaction turning points improves control.}
We next test whether concentrating on local outcome divergence improves
learning from a fixed deployment pool. Both variants draw from this pool
with equal data sizes and training budgets. Critical experience outperforms
full trajectories on Water Plant (90.7\% vs.\ 89.3\%) and Fold Glasses
(77.3\% vs.\ 74.0\%; Figure~\ref{fig:dewo_components}c).
The gains at matched budgets support the value of organizing supervision
around consequential interactions. Matched continuations highlight
different physical futures from comparable contexts, directing learning
toward the local changes that separate successful and failed execution.

\paragraph{Guidance benefits from selective activation.}
On Water Plant and Fold Glasses, value-gated guidance achieves 90.7\%
and 77.3\% success, compared with
86.7\% and 73.3\% without guidance and 78.0\% and 52.0\% under
continuous guidance (Figure~\ref{fig:dewo_components}d).
These gains support intervening selectively when estimated progress stalls.

\section{Related Work}
\label{sec:related}

\paragraph{World-action models for robot control.}
WAMs integrate visual dynamics modeling with action generation, using prediction of future observations to learn how actions affect the physical world~\citep{wang2026openwam,pai2025mimicvideo,zeng2024learning}.
A central architectural choice is how this predictive capability supports control:
predicted visual futures can serve as intermediate goals from which an inverse-dynamics model infers actions~\citep{tian2025predictive},
or future observations and actions can be generated jointly~\citep{ye2026world,kim2026cosmos}.
Predictive learning can also support control without explicitly generating futures at inference,
as demonstrated by FastWAM, which retains video prediction during training~\citep{yuan2026fastwam}.
Across these designs, the physical interactions available during training shape what the model can learn to predict.
Deployment brings new interactions that offer opportunities to refine this predictive capacity.
In this work, we explore how post-deployment training can refine this predictive capacity within existing WAM architectures to improve subsequent action generation.

\paragraph{Learning from deployment experience.}
Post-deployment learning commonly improves behavior through direct
action supervision or value-based reinforcement learning.
Expert corrections and interventions supply action
targets~\citep{ross2011reduction,liu2023robot}, while scalar rewards and
value estimates indicate which behaviors to
favor~\citep{wang2026learning}.
RECAP uses advantage-conditioned updates~\citep{intelligence2025pi}; DSRL optimizes a frozen
policy's latent noise~\citep{wagenmaker2025steering}.
These updates are especially challenging in dexterous manipulation,
where small changes in high-dimensional hand actions can disrupt
coordination and contact. LAMP addresses this difficulty with a learned hand-motion
prior that structures residual exploration around demonstrated
motions~\citep{yang2026lamp}.
WAMs provide a rich interaction prior through visual prediction,
learning how robot motions affect objects and how physical states
evolve~\citep{Finn2016}.
This predictive prior can itself be refined during deployment.
Both successful and failed attempts provide physical futures for
continued world modeling. Learning from these futures updates the
representations supporting action generation alongside
successful-action supervision.

\paragraph{Post-training world-action models.}
Recent work explores several forms of experience for post-training~\citep{bi2026motus2,hou2026worldmodelrobotlearning}.
RISE uses learned dynamics to supply imagined interactions for policy
improvement~\citep{yang2026rise}, while WAM post-training can also
update the predictive model itself.
WAM-RL uses successful online rollouts for video supervision alongside
reinforcement learning~\citep{qian2026wamrl}, whereas WAM-OPD obtains
teacher-generated video and action targets on histories visited by
the student~\citep{yang2026wam}.
Concurrent work FACT extends predictive supervision to failed rollouts,
retaining their action-conditioned video and task-progress targets
while masking action imitation~\citep{peng2026fact}.
These efforts investigate predictive learning within different training
objectives and data settings. Together, they motivate a unified view
of post-deployment adaptation through continued world modeling.
DEWO formalizes this direction as a post-deployment learning paradigm
for WAMs. It organizes learning around interaction turning points,
where matched successful and failed continuations reveal how similar
interactions evolve toward different physical futures.

\section{Conclusion}

Direct Experience World-Model Optimization (DEWO) makes continued world modeling an explicit part of post-deployment learning for WAMs, alongside action supervision. By organizing matched successful and failed continuations around interaction turning points, it turns their diverging visual futures into predictive supervision that supports control. Experiments show gains across three WAM formulations and two real-world robot embodiments. Ablations show that successful and failed visual experience improves prediction and control beyond successful-action supervision alone.

Our real-world evaluations show that DEWO makes existing spatial generalization more reliable. Looking ahead, a central question is whether learning around interaction turning points can provide a foundation for sustained self-improvement. We envision agents that trace how failures emerge, learn which changes in an interaction can lead to success, and apply this knowledge to subsequent attempts. Coupled with exploration that uncovers new informative interactions, this process could extend learning beyond the model’s initial competence.

\subsection*{AI use statement}

We used generative AI to assist with drafting, revising, and organizing the manuscript; identifying and comparing related work; discussing research framing and interpreting experimental results; and checking and revising mathematical formulations and derivations. The authors reviewed and edited the AI-assisted text, verified cited claims against the original sources, checked the mathematical derivations, and cross-checked reported results against the experimental records. We take responsibility for the final content of the paper, including all AI-assisted text, claims, and artifacts.

\bibliography{references}
\bibliographystyle{iclr2027_conference}

\appendix
\section{Experimental Protocols}
\label{app:experimental_protocols}

\subsection{Simulation Tasks and Evaluation Sets}
\label{app:simulation_protocol}
We evaluate Water Plant, Fold Glasses, Hammer Nail, Pick Bucket, and
Pinch Tongs from DexJoCo. Every simulation success evaluation, including
ablations, uses three 50-scene sets per task (seeds 0, 1, and 2) with one
fixed checkpoint. Task scores pool 150 trials; five-task averages weight
tasks equally and pool 750 trials. Seeds index evaluation scenes, not
training runs. Per-set results appear in Appendix~\ref{app:eval_rounds}.

\subsection{Initialization and Comparison Scope}
\label{app:comparison_protocol}
\label{app:model_initialization}
Table~\ref{tab:dewo_main}A compares SFT, RECAP, and DSRL on
$\pi_{0.5}$ and FastWAM, with DEWO included in the FastWAM comparison;
it also compares SFT and DEWO on FACT. Within each base-model comparison,
methods share the corresponding initial checkpoint. FACT uses the
FastWAM settings for collection, training, checkpoint selection, and
inference, with its own backbone and initial checkpoint.

For FACT and FastWAM, SFT follows FACT's second-stage training scheme
and uses all rollout data. Successful rollouts provide action-imitation,
video-prediction, and value/progress supervision. Failed rollouts retain
video-prediction and value/progress losses, while their action-imitation
loss is masked. For $\pi_{0.5}$, which has no video-prediction loss,
SFT uses successful rollouts.

RECAP uses value- and advantage-conditioned improvement, and DSRL
performs online reinforcement learning in the action policy's latent-noise
space. In the FastWAM comparison, SFT and RECAP use the same collected
deployment data as DEWO, with each method applying its own learning objective.

Table~\ref{tab:dewo_main}B varies prediction--action coupling in a shared
mixture-of-transformers (MoT) built from a Wan2.2 VideoDiT and an ActionDiT.
The three formulations are joint generation (Joint), video-then-action
inverse dynamics (IDM), and FastWAM. Each is compared before and after DEWO.
Repeated FastWAM entries refer to the same measurements as in panel A.
The optimization settings below describe five-task FastWAM adaptation
with DEWO and Joint WAM construction with DEWO-S.

\subsection{Collection Budgets and Training Data}
\label{app:budget_protocol}
Simulation collection uses one round, with seeds 10086--10135 and four
initial attempts per seed, or 200 initial attempts per task. At 30 environment
steps/s, the standard collection limit is 1,000 steps; Hammer Nail
collection uses the maximum expert-episode length plus three seconds.
Closed-loop evaluation uses a 1,000-step limit for all five tasks.

DEWO collection uses a query cap of 20 anchors with ten continuations
each, or at most 200 extra attempts \emph{per failed trajectory}.
Early stopping can reduce this cost. In the FastWAM comparison, SFT and
RECAP use the same collected data as DEWO; DSRL obtains experience through online
interaction, as described in Appendix~\ref{app:comparison_protocol}.

The four dataset roles are complete successful source trajectories
$D_0$, value-supervision scan entries $D_{\mathrm{scan}}$, retained
successful continuations or stitches $D_+$, and failed cliff windows
$D_{\mathrm{fail}}$ (the main text's $\mathcal D_-$). Table~\ref{tab:dewo_pool_counts} counts source entries before window expansion.

\begin{table}[htbp]
\centering\small
\caption{Five-task training pools before window expansion. Source-entry
counts differ from frame and interaction counts. The expert pool
contains 100 successful episodes per task.}
\label{tab:dewo_pool_counts}
\begin{tabularx}{\linewidth}{@{}Xrr@{}}
\toprule
Pool or source & DEWO & DEWO-S \\
\midrule
Expert episodes included in $D_0$ & 0 & 500 \\
Collected successful episodes included in $D_0$ & 106 & 106 \\
$D_0$ subtotal & 106 & 606 \\
$D_{\mathrm{scan}}$ & 1,396 & 1,396 \\
$D_+$ & 125 & 125 \\
$D_{\mathrm{fail}}$ & 134 & 134 \\
\midrule
Total source entries & 1,761 & 2,261 \\
\bottomrule
\end{tabularx}
\end{table}

DEWO uses deployment experience, while DEWO-S additionally includes
500 expert demonstrations. After window construction, samples from all
data roles are pooled and uniformly shuffled without rebalancing across
roles. Complete trajectories therefore contribute in proportion to their
number of windows; $D_{\mathrm{scan}}$ supplies only value targets.

\paragraph{Window construction and validation partition.}
FastWAM windows span 33 action-clock frames: one input and 32 future
frames/actions. Sampling video every four steps gives one input and
eight future frames. Complete $D_0$ and expert episodes use stride one.

For a collection seed with four successful initial attempts, one
complete successful rollout enters collected
$D_0$, and the remaining eligible successful rollouts enter validation.
Eligible complete successes from the other seeds also enter validation.
Scan entries and selected clips from failed source rollouts enter training;
expert episodes, when included, enter training only. The resulting
validation set contains 557 complete successful episodes. Training and
validation are split by episode before window expansion, with no shared
episodes; collection seeds may be shared across the two sets. The
validation set is used for open-loop evaluation and checkpoint selection.

Training ablations independently adapt the same Round~0 FastWAM checkpoint
on rollout-0 data with matched budgets
(Appendix~\ref{app:ablation_protocol}).

\subsection{Optimization and Checkpoint Selection}
\label{app:optimization_protocol}
Table~\ref{tab:run_settings} summarizes the two optimization settings.
Both use four GPUs with four examples per GPU in each forward/backward pass.

\begin{table}[htbp]
\centering\small
\caption{Optimization settings for five-task FastWAM adaptation with
DEWO and Joint WAM construction with DEWO-S.}
\label{tab:pending_records}
\label{tab:run_settings}
\renewcommand{\arraystretch}{1.12}
\begin{tabularx}{\linewidth}{@{}>{\raggedright\arraybackslash}p{.22\linewidth}>{\raggedright\arraybackslash}X>{\raggedright\arraybackslash}X@{}}
\toprule
Setting & DEWO & DEWO-S \\
\midrule
Initialization & Five-task FastWAM checkpoint at 55,000 updates &
Pretrained VideoDiT and ActionDiT components \\
Trainable parameters & Text cross-attention K/V residual adapters and
value head & Full video/action DiTs and MoT, proprioception/outcome
encoders, and value head \\
Frozen components & MoT, VideoDiT, ActionDiT backbone weights &
No additional freezing \\
Adapter & Rank 16, scale parameter 16; video and action experts &
None \\
Maximum updates & 10,000 & 100,000 \\
Checkpoint save interval & 2,500 updates & 5,000 updates \\
Selected checkpoint & After 10,000 adaptation updates & After 35,000 training updates \\
Validation protocol & Open-loop evaluation on the validation split &
Open-loop evaluation on the validation split \\
Selection basis & Open-loop validation loss & Open-loop validation loss \\
Simulation CFG scale & $w=0.2$, value-gated & $w=0$ \\
\bottomrule
\end{tabularx}
\end{table}

\paragraph{DEWO initialization and adaptation.}
DEWO adapts the five-task FastWAM checkpoint at 55,000 updates.
The backbone is frozen, and only low-rank residual adapters in the
key and value projections of text cross-attention and the value head
are optimized. Both parameter groups use a learning rate of $10^{-4}$.
The selected checkpoint is obtained after 10,000 adaptation updates.

\paragraph{DEWO-S initialization and selection.}
DEWO-S initializes VideoDiT with Wan2.2-TI2V-5B and uses a pretrained
ActionDiT to construct a Joint WAM. The full model is optimized under
the DEWO objective. All five tasks are evaluated using the same
checkpoint, selected after 35,000 training updates.

For both runs, checkpoints are selected using open-loop evaluation
loss on the validation split described in
Appendix~\ref{app:budget_protocol}. The $3\times50$ closed-loop
evaluation sets are used to report task success, not to select
checkpoints.

\begin{table}[htbp]
\centering\small
\caption{Collection, temporal, and deployment settings. The guidance
interval includes replans 10--24 for all five FastWAM DEWO simulation tasks.}
\label{tab:confirmed_settings}
\begin{tabularx}{\linewidth}{@{}p{.49\linewidth}X@{}}
\toprule
Setting & Value \\
\midrule
Simulation evaluation & 3 sets $\times$ 50 scenes per task \\
Initial simulation collection & 50 seeds $\times$ 4 attempts per task \\
Simulation / real continuations per anchor & $K=10$ / $K=4$ \\
Simulation query start / interval & Step 96 / 24 steps \\
Maximum queries per failed trajectory & 20 anchors \\
Intermediate recoverability-drop threshold & At least $4/10$ \\
Training window / stride & 33 action-clock frames / 1 \\
Predicted / executed action horizon & 32 / 24 actions \\
Simulation environment frequency & 30 steps/s \\
Temporal discount $\gamma$ & 0.99 \\
Value-stall threshold / horizon & $\alpha=1.27$ / $H=5$ replans \\
Guidance interval (inclusive) & Replans 10--24 \\
Real-world grid / spacing & $3\times3$ / approximately 10\,cm \\
Real-world trials per cell and checkpoint & 10 \\
Real-world deployment checkpoints & R0, R1, R2 \\
\bottomrule
\end{tabularx}
\end{table}
\FloatBarrier

\section{DEWO Implementation Details}
\label{app:dewo_implementation}

\subsection{Critical-Experience Queries}
\label{app:critical_experience}
The continuation policy is the deployed checkpoint used for collection.
For a failed simulation rollout, queries begin at step 96 and advance
along the original trajectory in steps of 24. At each anchor, ten sampled
continuations estimate recoverability as $p_i=k_i/10$. Here $i$ indexes
queried anchors, whose environment times satisfy $t_{i+1}-t_i=24$.
This fraction is used to select experience by comparing neighboring
queried anchors (Eq.~\ref{eq:critical_event}). The discounted value targets
used for progress monitoring are defined in Appendix~\ref{app:value_guidance}.

\paragraph{Stopping and terminal-boundary retention.}
At the first anchor with $0/10$ successful continuations, exploration of
all later positions on that original trajectory stops. The zero-success
query and its failed candidates are used as a stopping signal and are
not retained as training entries. At the immediately preceding queried
anchor with nonzero success, exactly one successful continuation is
retained: the shortest successful candidate. The original failed-rollout
window starting at that same anchor supplies the failed counterpart.
If there is no preceding nonzero anchor, this rule yields no successful
continuation. A failed trajectory also stops being queried when the
20-anchor cap is reached.

\paragraph{Intermediate drops.}
When two neighboring queried anchors satisfy
\begin{equation}
p_i-p_{i+1}\geq 0.4,
\label{eq:dewo_drop_selection}
\end{equation}
the original-rollout window beginning at the earlier anchor $i$ is a
failed cliff example. Equality is included: a drop from $9/10$ to
$5/10$ labels the segment beginning at the $9/10$ node.
This rule retains no additional successful continuation. Selected failed
windows form $D_{\mathrm{fail}}$ and receive the failure condition.
Scan entries in $D_{\mathrm{scan}}$ supply value targets only.

\paragraph{Physical queries.}
For real robots, candidate anchors are proposed by inspecting failed
rollouts. In the grasp-and-place tasks, this diagnosis traces failures to
pre-grasp alignment, contact, or an unstable grasp. The scene is restored
to a corresponding pre-grasp configuration, and $K=4$ continuations are
sampled per anchor. Physical restoration matches task-relevant scene and
robot state; it does not reproduce identical camera frames or the exact
observation prefix. The successful and failed alternatives supply
training examples without a pairwise ranking or contrastive objective.

\subsection{Outcome Conditions, Pool Mixing, and Losses}
\label{app:loss_and_replay}
The base condition $c_{\mathrm{base}}$ is the task instruction, e.g.,
\texttt{Fold the glasses and place them into the case.} for Fold Glasses.
Success and failure conditions append
\texttt{Successful execution.} and \texttt{Failed execution.}, respectively.

$D_0$ and $D_{\mathrm{scan}}$ use the base condition, while
$D_{\mathrm{fail}}$ uses the failure condition. For $D_+$, the success
condition is replaced by the base instruction with probability 0.1.

\begin{table}[htbp]
\centering\small
\caption{Per-role supervision weights. Action and value columns apply
to both methods. Value is supervised separately; failed action targets are masked.}
\label{tab:dewo_role_losses}
\begin{tabularx}{\linewidth}{@{}Xrrrr@{}}
\toprule
Role & Video (DEWO) & Video (DEWO-S) & Action & Value \\
\midrule
$D_0$ & 1 & 1 & 1 & 1 \\
$D_+$ & 1 & 1 & 1 & 1 \\
$D_{\mathrm{scan}}$ & 0 & 0 & 0 & 1 \\
$D_{\mathrm{fail}}$ & 1 & 1 & 0 & 1 \\
\bottomrule
\end{tabularx}
\end{table}

Both settings retain the native video and action prediction objectives.
Table~\ref{tab:dewo_role_losses} gives the supervision weights for each
data role. Successful trajectories and continuations supply video,
action, and value supervision; failed windows supply video and value
supervision, with their action targets masked.

Writing $\overline{\mathcal L}_{v}$,
$\overline{\mathcal L}_{a}$, and $\overline{\mathcal L}_{u}$ for the
video-prediction, action-prediction, and value-regression losses,
respectively, after their role weights and masks, the training objectives
are
\begin{align}
\mathcal L_{\mathrm{DEWO}}
&=\overline{\mathcal L}_{v}+\overline{\mathcal L}_{a}
 +\overline{\mathcal L}_{u}
 +0.1\,\mathcal L_{\mathrm{identity}}
 +0.05\,\mathcal L_{\mathrm{action\mbox{-}residual}},
\label{eq:dewo_recorded_loss}\\
\mathcal L_{\mathrm{DEWO\mbox{-}S}}
&=\overline{\mathcal L}_{v}+\overline{\mathcal L}_{a}
 +\overline{\mathcal L}_{u}.
\label{eq:dewos_recorded_loss}
\end{align}
For adapter-based DEWO, $\mathcal L_{\mathrm{identity}}$ penalizes the
video residual and $\mathcal L_{\mathrm{action\mbox{-}residual}}$
penalizes the action residual, with coefficients 0.1 and 0.05.
DEWO-S optimizes the full model without these adapter regularizers.
Value targets for all data roles follow the return construction below.

In the FastWAM and Joint implementations described here, visual
prediction uses no explicit executed-action input. Visual supervision
affects control through the architecture's native coupling, updating
adapters in DEWO and the backbone in DEWO-S. During inference, action
guidance combines the base and success conditions, while visual
prediction remains base-conditioned.

Checkpoint selection uses the mean training objective on the validation
set, including the video, action, and value terms and, for DEWO, the
residual regularizers. The prediction metric in
Appendix~\ref{app:prediction_metric} evaluates only the video term.

\subsection{Value Learning and Video Features}
\label{app:value_guidance}
The recoverability estimates in Section~\ref{sec:critical_experience} select
training experience. For progress monitoring during execution, the value
head $V_\phi(h_t)$ predicts discounted terminal success from the WAM's video
representations. It reads current-observation VideoDiT tokens with feature
dimension 3,072. For this encoding the adapter
is disabled and the task uses base text. The tokens are detached
(\emph{stop-gradient}), so the value loss does not backpropagate into
the main DiT. Mean pooling over the token sequence produces a
$[B,3072]$ representation, followed by LayerNorm and an MLP:
\begin{equation}
3072 \;\longrightarrow\; 256 \;\longrightarrow\; \mathrm{GELU}
\;\longrightarrow\; 256 \;\longrightarrow\; \mathrm{GELU}
\;\longrightarrow\; 1.
\end{equation}
Clamping the logit to $[-16,16]$ and applying a sigmoid gives $V(h)\in(0,1)$.

For a continuation $\tau$ with final decision step $T-1$, the return is
$G_t(\tau)=\gamma^{T-1-t}Y(\tau)$ with $\gamma=0.99$, where $Y(\tau)$ is the
binary success indicator. Failed continuations contribute zero, while
successful continuations contribute larger targets when completion is closer.
At a scan anchor, the regression target averages the returns from its $M_t$
\emph{extra} continuations $\tau_{t,i}$:
\begin{equation}
\widehat V_t
=\frac{1}{M_t}\sum_{i=1}^{M_t}\gamma^{T_i-1-t}Y(\tau_{t,i})
=\frac{1}{M_t}\sum_{i=1}^{M_t}G_t(\tau_{t,i}),
\label{eq:sampled_progress_target}
\end{equation}
where $T_i-1$ is the final decision step of continuation $i$.
The source branch is excluded. Each extra continuation contributes
once to this average, including any continuation subsequently retained
in $D_+$.

For a retained success $\tau^+$ with final decision step $T_+-1$,
the return at step $u$ is
\begin{equation}
G_u(\tau^+)=\gamma^{T_+-1-u}Y(\tau^+)
=\gamma^{T_+-1-u}.
\label{eq:dewo_positive_continuation_return}
\end{equation}
At queried contexts, the continuation average takes precedence over
the return of an individual retained branch. Intervening contexts use
trajectory returns or discounted temporal backup. The value objective
$\mathcal L_u$ applies Huber (SmoothL1) regression between $V(h_t)$ and
its target, with unit weight for all data roles
(Table~\ref{tab:dewo_role_losses}).

The value head shares the optimizer and training batches with the
adapter in DEWO or the full model in DEWO-S, while its input features
remain detached. At inference, progress is estimated from the current
observation using the same base-conditioned encoding, with adapters
disabled; the value head requires no future-video generation.

\subsection{Discount-Informed Progress Gating}
\label{app:dewo_discount_gate}
Monitoring progress over a window gives execution time to advance before
intervention. Let $V_j$ denote the value estimate at replan $j$. For $j\geq H$,
the stall indicator is
\begin{equation}
z_j
=\mathbb{I}\!\left[
\max_{1\leq k\leq H}V_{j-H+k}<\alpha V_{j-H}
\right].
\label{eq:progress_stall_indicator}
\end{equation}
Here, $H$ is the monitoring horizon and $\alpha$ the required value-increase
factor. Thus, a stall is detected when no estimate within the window reaches
this factor relative to its starting value. The deployed gate $g_j$ follows
$z_j$ within the eligible replanning interval and is zero outside it.

Along a fixed successful continuation, the target satisfies
\begin{equation}
G_t=\gamma^{T-1-t},\qquad
\frac{G_{t+\Delta}}{G_t}=\gamma^{-\Delta}.
\label{eq:discount_reference}
\end{equation}
At $\gamma=0.99$, a reference interval of 24 discount steps gives
$0.99^{-24}\simeq1.2728$, motivating the relative threshold
$\alpha=1.27$. We monitor this increase over $H=5$ replans.
The discount reference sets the required increase, while $H$ sets
the time allowed to achieve it.

The model predicts 32 actions and executes 24 before replanning.
At 30 simulation steps/s, one executed chunk spans 0.8\,s, so the
monitoring window covers 4.0\,s of simulated execution.
All five FastWAM DEWO simulation tasks use this rule with the same
$\alpha$ and $H$. Guidance follows the stall indicator from replan 10
through 24, inclusive, and is disabled outside this interval.

The action prediction combines the base and success branches as
\begin{equation}
f^{\mathrm{act}}_{\mathrm{guided},s}
=f^{\mathrm{act}}_{b,s}
+g_j\,w\bigl(f^{\mathrm{act}}_{+,s}-f^{\mathrm{act}}_{b,s}\bigr),
\qquad w=0.2.
\label{eq:dewo_recorded_cfg}
\end{equation}
The scale is shared across the five FastWAM tasks. Active guidance
therefore interpolates the two predictions as
$0.8f^{\mathrm{act}}_{b,s}+0.2f^{\mathrm{act}}_{+,s}$.
DEWO-S simulation evaluations use $w=0$. The relative progress rule
is a heuristic applied to estimated values, with no low-value cutoff.

\subsection{DEWO-S: Construction from Pretrained Components}
\label{app:dewos_protocol}
DEWO-S constructs the Joint WAM from the pretrained VideoDiT and ActionDiT
components described in Appendix~\ref{app:optimization_protocol}.
The DEWO objective trains the full video/action/MoT backbone,
proprioception and outcome encoders, and value head. The value head's
final layer is initialized to zero. This applies experience-based
supervision during WAM construction, before subsequent deployment
adaptation. Table~\ref{tab:dewos_rounds} reports the simulation results
by evaluation set. The real-world Round~0 model follows the same
construction procedure with its own checkpoint.

\FloatBarrier

\section{Real-World Evaluation and Deployment}
\label{app:real_protocol}

\subsection{Platforms, Tasks, and Success Criteria}
\label{app:real_hardware}
\label{app:real_trial_protocol}
The two embodiments pair the Xingchen platform with the Wuji hand and
the Tianji platform with the Sharpa hand. Each is evaluated on a water
bottle, tape, an eraser, and a tennis ball. A trial succeeds only when
the robot grasps the object and places it in the target basket.
The reported counts are complete-task successes; grasp acquisition
and stability are the principal difficulties identified by inspection.

Each object--embodiment pair uses a fixed $3\times3$ initial-position
grid with approximately 10\,cm between neighboring grid points and
ten trials per cell. This gives 90 trials per pair, 360 per hand, and
720 per checkpoint. Initial training data are concentrated around
the center region. All nine cells enter the full-grid metric, including
locations with no observed successes.

\subsection{Visual Observations and Preprocessing}
\label{app:camera_inputs}
The real-world policy takes a head-view and a wrist-view RGB image.
Simulation observations use front and wrist RGB images at
$640\times640$ pixels each. In both settings, each view is resized to
$224\times224$, and the two images are concatenated horizontally to
form an input of height 224 and width 448. Real-world videos are stored
at 30 frames/s.

\subsection{Round Definitions and Continuations}
\label{app:real_rounds}
Round~0 is the initial real-world DEWO-S checkpoint. Rounds~1 and~2
follow one and two additional collection--optimization cycles,
respectively. These checkpoints are denoted R0, R1, and R2,
corresponding to Round~0, Round~1, and Round~2. Three checkpoints give
2,160 evaluation
trials. The separate $\pi_{0.5}$ reference contributes another 720
trials and is not a DEWO-S deployment round.
Human-assisted selection, matched physical restoration, and four
continuations per anchor follow Appendix~\ref{app:critical_experience}.
These physical deployment rounds are distinct from the single
simulation collection round. Spatial groups are defined from initial
evaluation outcomes as described below.

\subsection{Fixed Spatial Groups and Aggregation}
\label{app:spatial_aggregation}
For descriptive analysis, all cells are grouped by whether their Round~0
evaluation has at least one success. Membership is fixed thereafter.
The R0-nonzero group includes centers and contains 14 Wuji cells and 15
Sharpa cells. Its denominator is 140 and 150 trials per round,
respectively, or 290 pooled. We further split this group into center
and initially nonzero off-center cells to describe spatial structure.
Across eight object--embodiment pairs, there are eight center cells,
21 initially nonzero off-center cells, and 43 initially zero off-center
cells. For group $S$ and round $r$, the success rate is
\begin{equation}
\mathrm{SR}_r(S)=\frac{\sum_{u\in S} n_{r,u}}{10|S|}\times100\%,
\end{equation}
where $n_{r,u}$ is the number of successes at cell $u$. The full-grid rate
uses all 72 cells. These groups describe observed initial competence,
with full-grid success reported separately over all evaluation locations.

\begin{table}[htbp]
\centering\small
\caption{Pooled spatial success counts and fixed denominators. The initially
nonzero subtotal combines the center and nonzero off-center rows.}
\label{tab:spatial_denominators}
\begin{tabular*}{\linewidth}{@{\extracolsep{\fill}}lrrrr@{}}
\toprule
Spatial group & Trials/round & R0 & R1 & R2 \\
\midrule
Center & 80 & 64 & 71 & 71 \\
Initially nonzero off-center & 210 & 84 & 138 & 137 \\
Initially zero off-center & 430 & 0 & 0 & 0 \\
\midrule
All initially nonzero (including center) & 290 & 148 & 209 & 208 \\
All cells & 720 & 148 & 209 & 208 \\
\bottomrule
\end{tabular*}
\end{table}
Among the 29 cells with a nonzero initial count, 23 improve, five are
unchanged, and one decreases from R0 to R2. None of the 43 initially zero
cells records a success in the subsequent evaluations. The group-wise
trends are descriptive because membership is defined using R0 evaluation
outcomes; the full spatial maps and pooled success rates report
performance over all locations.
\FloatBarrier

\section{Ablation Protocols and Video Metric}
\label{app:ablation_protocol}
\label{app:ablation_controls}
The ablations use FastWAM on Water Plant and Fold Glasses, with
$3\times50$ evaluation trials per task and configuration. Training
variants independently adapt the same Round~0 checkpoint on rollout-0
source data, with matched training-data quantities and budgets.
Controls include the initial model, action-only adaptation, and unguided inference.

\subsection{Visual and Action Supervision}
\label{app:supervision_ablation}
$A$ uses successful-action supervision. $A+V^+$ adds successful visual
futures, and $A+V^++V^-$ additionally uses failed visual futures.
Successful visual supervision comes from $D_0$ and $D_+$; failed
visual supervision comes from $D_{\mathrm{fail}}$, with failed action
targets masked. Figure~\ref{fig:dewo_components}a,b reports control
and prediction outcomes for these variants.

\subsection{Offline Validation Video Loss}
\label{app:prediction_metric}
The prediction metric is the native video prediction loss evaluated
offline on the validation set, using the model's loss normalization,
masks, and generative-time/noise handling. It measures the video term
alone; checkpoint selection uses the total validation objective
(Appendix~\ref{app:loss_and_replay}).

Lower values are better. Comparing $A$ with the full supervision
ablation reduces video loss from 0.196 to 0.132 on Water Plant and
from 0.214 to 0.176 on Fold Glasses, relative reductions of 32.7\%
and 17.8\%. Lower validation video loss accompanies higher task success
on both tasks.

\subsection{Experience Selection and Guidance Timing}
\label{app:selection_ablation}
\label{app:guidance_ablation}
The selection ablation contrasts full-trajectory training with
critical experience at matched training-data quantities and budgets.
Both adapt the same initial model using the shared rollout-0 source pool.
The guidance comparison evaluates no guidance, continuous guidance,
and value-gated guidance on the same adapted FastWAM checkpoint, varying
only inference settings (Figure~\ref{fig:dewo_components}d).

\section{Reference-Guided Action Generation}
\label{app:dewo_reference_guidance}

\paragraph{Setup.}
Fix the decision context $h$ and any other conditions shared by the two
action branches. We write $q_+(a\mid h)$ for the critical-success action
distribution and $q_b(a\mid h)$ for the base reference. These are idealized
distributions associated with the respective training conditions. The base
is not assumed to be the marginal distribution of all successful and failed
deployment actions. Consequently, the density ratio below is not identified
with the true environment probability of task success.

\paragraph{A reference-regularized preference objective.}
Assume that both densities are positive on the same support and vanish
outside it. Define $R(a;h)=\log[q_+(a\mid h)/q_b(a\mid h)]$ there.
For $\lambda\ge0$, consider the functional
\begin{equation}
\mathcal J_\lambda(q)
=\lambda\,\mathbb E_{a\sim q}[R(a;h)]
-D_{\mathrm{KL}}(q\Vert q_b).
\label{eq:dewo_reference_functional}
\end{equation}
Assume
\begin{equation}
0<Z_\lambda(h)
=\int q_b(a\mid h)\exp\{\lambda R(a;h)\}\,da<\infty.
\end{equation}
Over distributions for which the functional is well-defined, the maximizing
distribution is
\begin{equation}
q_\lambda^*(a\mid h)
=\frac{q_b(a\mid h)\exp\{\lambda R(a;h)\}}{Z_\lambda(h)}
=\frac{q_b(a\mid h)^{1-\lambda}q_+(a\mid h)^\lambda}
{Z_\lambda(h)}.
\label{eq:dewo_reference_tilt}
\end{equation}
Indeed, substitution gives
\begin{equation}
\mathcal J_\lambda(q)
=\log Z_\lambda(h)-D_{\mathrm{KL}}(q\Vert q_\lambda^*),
\end{equation}
which proves the claim. Thus the reference and critical-success
distributions are recovered at $\lambda=0$ and $\lambda=1$, respectively;
$\lambda>1$ emphasizes their relative experience preference further.
This variational objective provides an interpretation of the guidance
rule; the model is trained with the losses in
Appendix~\ref{app:loss_and_replay}.

\paragraph{Local guidance along a shared Gaussian path.}
Let $q_{c,s}(x\mid h)$, $c\in\{b,+\}$, be the densities obtained by applying
the same Gaussian corruption path to the two action distributions.
At every interior generative time $s$, exact scores obey
\begin{equation}
\nabla_x\log q_{+,s}(x\mid h)-\nabla_x\log q_{b,s}(x\mid h)
=\nabla_x\log\frac{q_{+,s}(x\mid h)}{q_{b,s}(x\mid h)}.
\label{eq:dewo_local_ratio_score}
\end{equation}
In particular, their linear combination is the score of the time-local
tilt
\begin{equation}
\widetilde q_{\lambda,s}(x\mid h)
\propto q_{b,s}(x\mid h)^{1-\lambda}q_{+,s}(x\mid h)^\lambda.
\label{eq:dewo_local_tilt}
\end{equation}

For clarity, use a noise-to-data flow convention
\begin{equation}
x_s=s a+(1-s)\epsilon,
\qquad\epsilon\sim\mathcal N(0,I),\quad0<s<1.
\end{equation}
For either branch, $v_c(x,s)=\mathbb E[a-\epsilon\mid x_s=x,c,h]$.
Gaussian conditioning gives
\begin{equation}
\nabla_x\log q_{c,s}(x\mid h)
=-\frac{\mathbb E[\epsilon\mid x_s=x,c,h]}{1-s},
\end{equation}
and hence
\begin{equation}
v_c(x,s)=\frac{x}{s}
+\frac{1-s}{s}\nabla_x\log q_{c,s}(x\mid h).
\label{eq:dewo_velocity_score}
\end{equation}
It follows that
\begin{equation}
v_+(x,s)-v_b(x,s)
=\frac{1-s}{s}\nabla_x\log
\frac{q_{+,s}(x\mid h)}{q_{b,s}(x\mid h)}.
\label{eq:dewo_velocity_ratio}
\end{equation}
The general Gaussian-schedule relation is given by
\citet{zheng2023guidedflowsgenerativemodeling}. Here $s$ indexes generative time; it is
distinct from environment time. Implementations with the opposite time
convention must transform both the field and the integration direction.

With $\lambda=g_j w$ independent of the current generative state,
the guided field is
\begin{equation}
v_{\mathrm{guided}}=v_b+g_j w(v_+-v_b).
\end{equation}
The implemented network outputs approximate these fields. The interpretation
assumes a common probability path and shared non-outcome inputs between the
two branches. If visual latents serve as additional conditions, this statement
is conditional on those shared inputs, rather than a proof for arbitrary
joint multimodal sampling dynamics.

\paragraph{Scope of the interpretation.}
The equalities above describe a time-local guidance direction and a
reference-regularized preference. They do not imply that the complete guided
sampler exactly produces the clean-action density in
Eq.~\ref{eq:dewo_reference_tilt}. Time-local geometric mixtures need not form
the noising path of that endpoint distribution, a known limitation of a
literal endpoint-density interpretation of CFG.
They also do not establish monotonic improvement in true task success;
that claim must be assessed empirically.

\section{Simulation Results by Evaluation Set}
\label{app:eval_rounds}

Each seed identifies one fixed 50-scene evaluation set. The tables report
success percentages for those scenes and their arithmetic mean, rather than
variation across independently trained models. Table~\ref{tab:dewo_main}
reports the corresponding per-task means and equally weighted task averages.

\begin{table}[htbp]
\centering
\footnotesize
\caption{
Per-evaluation-set $\pi_{0.5}$ success rates (\%).
}
\label{tab:pi05_rounds}
\setlength{\tabcolsep}{5pt}
\renewcommand{\arraystretch}{1.04}
\begin{tabular*}{\linewidth}{@{\extracolsep{\fill}}llrrrr@{}}
\toprule
\textbf{Method} &
\textbf{Task} &
\textbf{Seed 0} &
\textbf{Seed 1} &
\textbf{Seed 2} &
\textbf{Mean} \\
\midrule
\multirow{5}{*}{Initial}
& Water Plant  & 76 & 70 & 74 & 73.3 \\
& Fold Glasses & 54 & 62 & 58 & 58.0 \\
& Hammer Nail  & 74 & 74 & 76 & 74.7 \\
& Pick Bucket  & 80 & 78 & 78 & 78.7 \\
& Pinch Tongs  & 66 & 64 & 58 & 62.7 \\
\midrule
\multirow{5}{*}{+ SFT}
& Water Plant  & 76 & 74 & 76 & 75.3 \\
& Fold Glasses & 58 & 58 & 62 & 59.3 \\
& Hammer Nail  & 76 & 78 & 76 & 76.7 \\
& Pick Bucket  & 86 & 86 & 84 & 85.3 \\
& Pinch Tongs  & 20 & 20 & 28 & 22.7 \\
\midrule
\multirow{5}{*}{+ RECAP}
& Water Plant  & 76 & 72 & 78 & 75.3 \\
& Fold Glasses & 52 & 48 & 54 & 51.3 \\
& Hammer Nail  & 80 & 84 & 76 & 80.0 \\
& Pick Bucket  & 84 & 84 & 84 & 84.0 \\
& Pinch Tongs  & 60 & 54 & 58 & 57.3 \\
\midrule
\multirow{5}{*}{+ DSRL}
& Water Plant  & 80 & 74 & 76 & 76.7 \\
& Fold Glasses & 58 & 52 & 54 & 54.7 \\
& Hammer Nail  & 84 & 76 & 84 & 81.3 \\
& Pick Bucket  & 86 & 86 & 86 & 86.0 \\
& Pinch Tongs  & 24 & 24 & 20 & 22.7 \\
\bottomrule
\end{tabular*}
\end{table}

\begin{table}[htbp]
\centering
\footnotesize
\caption{
Per-evaluation-set FastWAM success rates (\%).
}
\label{tab:fastwam_rounds}
\setlength{\tabcolsep}{5pt}
\renewcommand{\arraystretch}{1.04}
\begin{tabular*}{\linewidth}{@{\extracolsep{\fill}}llrrrr@{}}
\toprule
\textbf{Method} &
\textbf{Task} &
\textbf{Seed 0} &
\textbf{Seed 1} &
\textbf{Seed 2} &
\textbf{Mean} \\
\midrule
\multirow{5}{*}{Initial}
& Water Plant  & 88 & 86 & 92 & 88.7 \\
& Fold Glasses & 74 & 70 & 72 & 72.0 \\
& Hammer Nail  & 76 & 74 & 74 & 74.7 \\
& Pick Bucket  & 94 & 92 & 90 & 92.0 \\
& Pinch Tongs  & 78 & 72 & 78 & 76.0 \\
\midrule
\multirow{5}{*}{+ SFT}
& Water Plant  & 70 & 74 & 76 & 73.3 \\
& Fold Glasses & 74 & 78 & 72 & 74.7 \\
& Hammer Nail  & 76 & 80 & 80 & 78.7 \\
& Pick Bucket  & 84 & 86 & 84 & 84.7 \\
& Pinch Tongs  & 70 & 68 & 70 & 69.3 \\
\midrule
\multirow{5}{*}{+ RECAP}
& Water Plant  & 80 & 70 & 84 & 78.0 \\
& Fold Glasses & 70 & 74 & 74 & 72.7 \\
& Hammer Nail  & 72 & 70 & 72 & 71.3 \\
& Pick Bucket  & 90 & 90 & 92 & 90.7 \\
& Pinch Tongs  & 68 & 76 & 70 & 71.3 \\
\midrule
\multirow{5}{*}{+ DSRL}
& Water Plant  & 68 & 76 & 76 & 73.3 \\
& Fold Glasses & 80 & 76 & 78 & 78.0 \\
& Hammer Nail  & 72 & 76 & 76 & 74.7 \\
& Pick Bucket  & 88 & 94 & 90 & 90.7 \\
& Pinch Tongs  & 66 & 64 & 68 & 66.0 \\
\midrule
\multirow{5}{*}{+ DEWO}
& Water Plant  & 84 & 94 & 94 & 90.7 \\
& Fold Glasses & 78 & 66 & 88 & 77.3 \\
& Hammer Nail  & 80 & 64 & 74 & 72.7 \\
& Pick Bucket  & 94 & 96 & 90 & 93.3 \\
& Pinch Tongs  & 74 & 76 & 82 & 77.3 \\
\bottomrule
\end{tabular*}
\end{table}

\begin{table}[htbp]
\centering
\footnotesize
\caption{
Per-evaluation-set Joint WAM success rates (\%).
}
\label{tab:joint_rounds}
\setlength{\tabcolsep}{5pt}
\renewcommand{\arraystretch}{1.04}
\begin{tabular*}{\linewidth}{@{\extracolsep{\fill}}llrrrr@{}}
\toprule
\textbf{Method} &
\textbf{Task} &
\textbf{Seed 0} &
\textbf{Seed 1} &
\textbf{Seed 2} &
\textbf{Mean} \\
\midrule
\multirow{5}{*}{Initial}
& Water Plant  & 84 & 84 & 76 & 81.3 \\
& Fold Glasses & 78 & 70 & 76 & 74.7 \\
& Hammer Nail  & 92 & 78 & 76 & 82.0 \\
& Pick Bucket  & 96 & 86 & 96 & 92.7 \\
& Pinch Tongs  & 84 & 76 & 76 & 78.7 \\
\midrule
\multirow{5}{*}{+ DEWO}
& Water Plant  & 94 & 86 & 96 & 92.0 \\
& Fold Glasses & 74 & 70 & 82 & 75.3 \\
& Hammer Nail  & 86 & 80 & 72 & 79.3 \\
& Pick Bucket  & 98 & 94 & 92 & 94.7 \\
& Pinch Tongs  & 98 & 86 & 92 & 92.0 \\
\bottomrule
\end{tabular*}
\end{table}

\begin{table}[htbp]
\centering
\footnotesize
\caption{
Per-evaluation-set IDM WAM success rates (\%).
}
\label{tab:idm_rounds}
\setlength{\tabcolsep}{5pt}
\renewcommand{\arraystretch}{1.04}
\begin{tabular*}{\linewidth}{@{\extracolsep{\fill}}llrrrr@{}}
\toprule
\textbf{Method} &
\textbf{Task} &
\textbf{Seed 0} &
\textbf{Seed 1} &
\textbf{Seed 2} &
\textbf{Mean} \\
\midrule
\multirow{5}{*}{Initial}
& Water Plant  & 86 & 86 & 86 & 86.0 \\
& Fold Glasses & 66 & 66 & 72 & 68.0 \\
& Hammer Nail  & 88 & 88 & 80 & 85.3 \\
& Pick Bucket  & 86 & 90 & 82 & 86.0 \\
& Pinch Tongs  & 86 & 86 & 84 & 85.3 \\
\midrule
\multirow{5}{*}{+ DEWO}
& Water Plant  & 96 & 98 & 96 & 96.7 \\
& Fold Glasses & 70 & 70 & 76 & 72.0 \\
& Hammer Nail  & 82 & 82 & 80 & 81.3 \\
& Pick Bucket  & 90 & 88 & 88 & 88.7 \\
& Pinch Tongs  & 78 & 80 & 86 & 81.3 \\
\bottomrule
\end{tabular*}
\end{table}

\begin{table}[htbp]
\centering
\footnotesize
\caption{Per-evaluation-set DEWO-S success rates (\%). Parentheses give
successful trials / total trials. The final column reports the mean and
population standard deviation over the three 50-scene evaluation sets,
with pooled trial counts.}
\label{tab:dewos_rounds}
\setlength{\tabcolsep}{5pt}
\renewcommand{\arraystretch}{1.04}
\begin{tabular*}{\linewidth}{@{\extracolsep{\fill}}lrrrr@{}}
\toprule
\textbf{Task} & \textbf{Seed 0} & \textbf{Seed 1} & \textbf{Seed 2} & \textbf{Mean $\pm$ SD} \\
\midrule
Water Plant  & 94 (47/50) & 94 (47/50) & 96 (48/50) & $94.7 \pm 0.9$ (142/150) \\
Fold Glasses & 86 (43/50) & 82 (41/50) & 80 (40/50) & $82.7 \pm 2.5$ (124/150) \\
Hammer Nail  & 82 (41/50) & 74 (37/50) & 74 (37/50) & $76.7 \pm 3.8$ (115/150) \\
Pick Bucket  & 96 (48/50) & 96 (48/50) & 100 (50/50) & $97.3 \pm 1.9$ (146/150) \\
Pinch Tongs  & 100 (50/50) & 96 (48/50) & 98 (49/50) & $98.0 \pm 1.6$ (147/150) \\
\bottomrule
\end{tabular*}
\end{table}

\begin{table}[t]
\centering
\footnotesize
\caption{Per-evaluation-set FACT success rates (\%). Each method uses
one fixed checkpoint across three 50-scene evaluation sets per task.
Parentheses give successful trials / total trials. The final column
reports the mean over the three sets with pooled trial counts;
the all-task rows aggregate 250 trials per set and 750 trials overall.}
\label{tab:fact_eval_sets}
\setlength{\tabcolsep}{4pt}
\renewcommand{\arraystretch}{1.04}
\begin{tabular*}{\linewidth}{@{\extracolsep{\fill}}llrrrr@{}}
\toprule
\textbf{Method} & \textbf{Task} & \textbf{Eval. 0} & \textbf{Eval. 1} & \textbf{Eval. 2} & \textbf{Mean} \\
\midrule
\multirow{6}{*}{Initial}
& Water Plant & 68 (34/50) & 66 (33/50) & 64 (32/50) & 66.0 (99/150) \\
& Fold Glasses & 76 (38/50) & 64 (32/50) & 64 (32/50) & 68.0 (102/150) \\
& Hammer Nail & 26 (13/50) & 24 (12/50) & 26 (13/50) & 25.3 (38/150) \\
& Pick Bucket & 88 (44/50) & 84 (42/50) & 84 (42/50) & 85.3 (128/150) \\
& Pinch Tongs & 80 (40/50) & 76 (38/50) & 78 (39/50) & 78.0 (117/150) \\
& All five tasks & 67.6 (169/250) & 62.8 (157/250) & 63.2 (158/250) & 64.5 (484/750) \\
\midrule
\multirow{6}{*}{+ SFT}
& Water Plant & 76 (38/50) & 60 (30/50) & 70 (35/50) & 68.7 (103/150) \\
& Fold Glasses & 60 (30/50) & 62 (31/50) & 86 (43/50) & 69.3 (104/150) \\
& Hammer Nail & 40 (20/50) & 44 (22/50) & 40 (20/50) & 41.3 (62/150) \\
& Pick Bucket & 86 (43/50) & 86 (43/50) & 88 (44/50) & 86.7 (130/150) \\
& Pinch Tongs & 84 (42/50) & 84 (42/50) & 86 (43/50) & 84.7 (127/150) \\
& All five tasks & 69.2 (173/250) & 67.2 (168/250) & 74.0 (185/250) & 70.1 (526/750) \\
\midrule
\multirow{6}{*}{+ DEWO}
& Water Plant & 98 (49/50) & 94 (47/50) & 92 (46/50) & 94.7 (142/150) \\
& Fold Glasses & 76 (38/50) & 80 (40/50) & 80 (40/50) & 78.7 (118/150) \\
& Hammer Nail & 50 (25/50) & 42 (21/50) & 42 (21/50) & 44.7 (67/150) \\
& Pick Bucket & 94 (47/50) & 94 (47/50) & 90 (45/50) & 92.7 (139/150) \\
& Pinch Tongs & 94 (47/50) & 98 (49/50) & 94 (47/50) & 95.3 (143/150) \\
& All five tasks & 82.4 (206/250) & 81.6 (204/250) & 79.6 (199/250) & 81.2 (609/750) \\
\bottomrule
\end{tabular*}
\end{table}

\FloatBarrier

\section{Real-World Spatial Evaluation Grids}
\label{app:real_spatial}

Each entry counts complete grasp-and-place successes out of ten trials.
Grid rows are TL/TC/TR, ML/C/MR, and BL/BC/BR, from top-left to bottom-right;
C is the center. R0/R1/R2 denote Round~0/Round~1/Round~2.
To align these coordinates with Figure~\ref{fig:dewo_spatial}a, vertically
flip water-bottle matrices and transpose the other object matrices. This
display mapping preserves every count and center-cell membership.

\newcommand{\realgridtab}[1]{%
\setlength{\tabcolsep}{3.6pt}
\renewcommand{\arraystretch}{1.04}
\begin{tabular*}{\linewidth}{@{\extracolsep{\fill}}ll*{10}{r}@{}}
\toprule
\textbf{Emb.} &
\textbf{Object} &
\textbf{TL} & \textbf{TC} & \textbf{TR} &
\textbf{ML} & \textbf{C} & \textbf{MR} &
\textbf{BL} & \textbf{BC} & \textbf{BR} &
\textbf{Overall} \\
\midrule
#1
\bottomrule
\end{tabular*}
}

\begin{table}[p]
\centering
\footnotesize
\caption{Real-world spatial evaluation. Each spatial entry reports
complete grasp-and-place successes out of 10 trials; Overall reports
the total out of 90 trials and the corresponding success rate.
Panels A--D show $\pi_{0.5}$ and DEWO-S Rounds~0--2, respectively.}
\label{tab:real_grid_pi05}
\realgridtab{
\multicolumn{12}{@{}l}{\textit{A. $\pi_{0.5}$ reference}} \\
\addlinespace[2pt]
\multirow{4}{*}{Wuji}
& Water Bottle & 0 & 4 & 1 & 0 & 10 & 6 & 0 & 0 & 0 & 21/90 (23.3\%) \\
& Tape & 0 & 0 & 0 & 4 & 10 & 0 & 3 & 9 & 0 & 26/90 (28.9\%) \\
& Eraser & 0 & 0 & 0 & 2 & 5 & 0 & 1 & 3 & 0 & 11/90 (12.2\%) \\
& Tennis Ball & 0 & 0 & 0 & 0 & 4 & 0 & 0 & 2 & 0 & 6/90 (6.7\%) \\
\addlinespace[3pt]
\multirow{4}{*}{Sharpa}
& Water Bottle & 0 & 5 & 1 & 0 & 10 & 8 & 0 & 0 & 0 & 24/90 (26.7\%) \\
& Tape & 0 & 0 & 0 & 4 & 9 & 0 & 4 & 10 & 0 & 27/90 (30.0\%) \\
& Eraser & 0 & 0 & 0 & 3 & 6 & 0 & 2 & 5 & 0 & 16/90 (17.8\%) \\
& Tennis Ball & 0 & 0 & 0 & 1 & 9 & 0 & 0 & 1 & 0 & 11/90 (12.2\%) \\
\midrule
\multicolumn{12}{@{}l}{\textit{B. DEWO-S Round~0}
\label{tab:real_grid_r1}\label{tab:real_grid_dewo_r0}} \\
\addlinespace[2pt]
\multirow{4}{*}{Wuji}
& Water Bottle & 0 & 5 & 1 & 0 & 10 & 7 & 0 & 0 & 0 & 23/90 (25.6\%) \\
& Tape & 0 & 0 & 0 & 4 & 10 & 0 & 4 & 9 & 0 & 27/90 (30.0\%) \\
& Eraser & 0 & 0 & 0 & 2 & 6 & 0 & 1 & 3 & 0 & 12/90 (13.3\%) \\
& Tennis Ball & 0 & 0 & 0 & 0 & 4 & 0 & 0 & 2 & 0 & 6/90 (6.7\%) \\
\addlinespace[3pt]
\multirow{4}{*}{Sharpa}
& Water Bottle & 0 & 5 & 2 & 0 & 10 & 8 & 0 & 0 & 0 & 25/90 (27.8\%) \\
& Tape & 0 & 0 & 0 & 5 & 9 & 0 & 4 & 9 & 0 & 27/90 (30.0\%) \\
& Eraser & 0 & 0 & 0 & 3 & 6 & 0 & 2 & 5 & 0 & 16/90 (17.8\%) \\
& Tennis Ball & 0 & 0 & 0 & 2 & 9 & 0 & 0 & 1 & 0 & 12/90 (13.3\%) \\
\midrule
\multicolumn{12}{@{}l}{\textit{C. DEWO-S Round~1}
\label{tab:real_grid_r2}\label{tab:real_grid_dewo_r1}} \\
\addlinespace[2pt]
\multirow{4}{*}{Wuji}
& Water Bottle & 0 & 8 & 5 & 0 & 10 & 9 & 0 & 0 & 0 & 32/90 (35.6\%) \\
& Tape & 0 & 0 & 0 & 7 & 10 & 0 & 6 & 9 & 0 & 32/90 (35.6\%) \\
& Eraser & 0 & 0 & 0 & 5 & 7 & 0 & 4 & 6 & 0 & 22/90 (24.4\%) \\
& Tennis Ball & 0 & 0 & 0 & 0 & 6 & 0 & 0 & 5 & 0 & 11/90 (12.2\%) \\
\addlinespace[3pt]
\multirow{4}{*}{Sharpa}
& Water Bottle & 0 & 8 & 5 & 0 & 10 & 9 & 0 & 0 & 0 & 32/90 (35.6\%) \\
& Tape & 0 & 0 & 0 & 7 & 10 & 0 & 7 & 9 & 0 & 33/90 (36.7\%) \\
& Eraser & 0 & 0 & 0 & 6 & 8 & 0 & 5 & 7 & 0 & 26/90 (28.9\%) \\
& Tennis Ball & 0 & 0 & 0 & 6 & 10 & 0 & 0 & 5 & 0 & 21/90 (23.3\%) \\
\midrule
\multicolumn{12}{@{}l}{\textit{D. DEWO-S Round~2}
\label{tab:real_grid_r3}\label{tab:real_grid_dewo_r2}} \\
\addlinespace[2pt]
\multirow{4}{*}{Wuji}
& Water Bottle & 0 & 7 & 6 & 0 & 10 & 9 & 0 & 0 & 0 & 32/90 (35.6\%) \\
& Tape & 0 & 0 & 0 & 8 & 10 & 0 & 5 & 9 & 0 & 32/90 (35.6\%) \\
& Eraser & 0 & 0 & 0 & 5 & 7 & 0 & 5 & 5 & 0 & 22/90 (24.4\%) \\
& Tennis Ball & 0 & 0 & 0 & 0 & 6 & 0 & 0 & 4 & 0 & 10/90 (11.1\%) \\
\addlinespace[3pt]
\multirow{4}{*}{Sharpa}
& Water Bottle & 0 & 8 & 6 & 0 & 10 & 8 & 0 & 0 & 0 & 32/90 (35.6\%) \\
& Tape & 0 & 0 & 0 & 7 & 10 & 0 & 8 & 8 & 0 & 33/90 (36.7\%) \\
& Eraser & 0 & 0 & 0 & 6 & 8 & 0 & 6 & 6 & 0 & 26/90 (28.9\%) \\
& Tennis Ball & 0 & 0 & 0 & 5 & 10 & 0 & 0 & 6 & 0 & 21/90 (23.3\%) \\
}
\end{table}

\end{document}